\documentclass[10pt,twocolumn,letterpaper]{article}

\usepackage{cvpr}
\usepackage{times}
\usepackage{epsfig}
\usepackage{graphicx}
\usepackage{amsmath}
\usepackage{amssymb}
\usepackage{mathtools}
\usepackage{algorithm}
\usepackage[noend]{algpseudocode}
\usepackage{dpr}
\usepackage{fec}
\usepackage{makecell}
\usepackage{subcaption}
\usepackage{booktabs}
\usepackage{enumitem}
\usepackage{comment}
\usepackage{xcolor, colortbl}
\usepackage{amsfonts,dcolumn}
\usepackage{caption}
\usepackage[pagebackref=true,breaklinks=true,letterpaper=true,colorlinks,bookmarks=false]{hyperref}

\cvprfinalcopy % *** Uncomment this line for the final submission

\def\cvprPaperID{1684} % *** Enter the CVPR Paper ID here

\newcommand{\liblinear}{Liblinear{}}
\newcommand{\openmp}{OpenMP{}}
\newcommand{\vgg}{VGG16{}}
\newcommand{\yolo}{YOLO{}}

\newcommand{\pytorch}{PyTorch{}}
\newcommand{\algname}{\text{Fast Sparse Convolutional Neural Network}{}}
\newcommand{\algacro}{\text{FSCNN}{}}
\newcommand{\cnn}{CNN}

\newcommand{\sialg}{Sparse Input CNN{}}

\newcommand{\csrc}{\text{CSR(C)}}
\newcommand{\sfalg}{Sparse Filter CNN{}}

\newcommand{\escoin}{\text{Escoin}}
\newcommand{\eie}{\text{EIE}}
\newcommand{\scnn}{\text{DSCNN}}

\newcommand{\myeq}{\stackrel{\mathclap{\normalfont\mbox{\small def}}}{=}}

\def \R {\mathbb{R}}

\def \S {\mathbb{S}}

\definecolor{Gray}{gray}{0.85}

\ifcvprfinal\fi
\begin{document}
	
	%%%%%%%%% TITLE
	\title{\algacro{}: A Fast Sparse Convolution Neural Network Inference System}	
	\author{
		Bo Ji\\
		Zhejiang University\\
		Zhejiang, China\\		
		{\tt\small jibo27@zju.edu.cn}
		% For a paper whose authors are all at the same institution,
		% omit the following lines up until the closing ``}''.
		% Additional authors and addresses can be added with ``\and'',
		% just like the second author.
		% To save space, use either the email address or home page, not both
		\and
		Tianyi Chen\\
		Microsoft\\
		Redmond, USA\\
		{\tt\small Tianyi.Chen@microsoft.com}
	}
	\maketitle
	%\thispagestyle{empty}
	
	%%%%%%%%% ABSTRACT
	\begin{abstract}
Convolution neural networks (CNNs) have achieved remarkable success, but typically accompany high computation cost and numerous redundant weight parameters. To reduce the FLOPs, structure pruning is a popular approach to remove the entire hidden structures via introducing coarse-grained sparsity. Meanwhile, plentiful pruning works leverage fine-grained sparsity instead (sparsity are randomly distributed), whereas their sparse models lack special designed computing library for potential speedup. In this technical report, we study and present an efficient convolution neural network inference system to accelerate its forward pass by utilizing the fine-grained sparsity of compressed CNNs. Our developed \algacro{} is established based on a set of specialized designed sparse data structures, operators and associated algorithms. Experimentally, we validate that \algacro{} outperforms standard deep learning library~\pytorch{} on popular CNN architectures such as \vgg{} if sufficiently high sparsity exhibits. However, due to the contiguity issue of sparse operators, \algacro{} is typically not comparable with highly optimized dense operator. Therefore, coarse-grained (structured) sparsity is our recommendation for generic model compression. \footnote{\textbf{Remark here that this technical report was written on 2019. We recently released it for sharing the implementation details for the sparse operators that perform on 3D or 4D tensors.}}
	\end{abstract}
	
	%\include{introduction/introduction}

	%%%%%%%%% BODY TEXT
	\section{Introduction}\label{sec.introduction}
	%-------------------------------------------------------------------------	
	Convolutional Neural Network (CNN) has demonstrated its success in plentiful computer vision application~\cite{krizhevsky2012imagenet,girshick2014rich, redmon2017yolo9000}. However, the structure of CNN typically results in high computational cost during inference, which encumbers its deployment into mobile devices~\cite{lane2016deepx}. To address this issue, extensive works have been done on accelerating neural networks by compression~\cite{han2015deep, jaderberg2014speeding, he2017channel, cheng2017survey},  quantization~\cite{rastegari2016xnor},  tensor factorization~\cite{jaderberg2014speeding} and special convolution implementations~\cite{chollet2017xception, howard2017mobilenets, wu2018shift}.% to decomposes weights into light-weight pieces.

	As a common practice, model compression is achieved by reducing the number of parameters and computation of each layer in deep neural networks. It can be largely evolved into two categories: (i) fine-grained pruning and (ii) structured pruning~\cite{anwar2017structured}. Fine-grained pruning aims to prune individual unimportant elements in weight tensors to zeros, which achieves very high sparse weights with no loss of accuracy~\cite{han2015learning}. To achieve sparse weights, sparsity-inducing regularized optimization, \eg $\ell_1$-regularization, is the common technique that has been demonstrated to yield extremely high sparse solution with satisfying model accuracy effectively~\cite{liu2015sparse}. More specifically, some state-of-the-art stochastic $\ell_1$-regularized solvers~\cite{jia2018irda} are able to compute model weights with sparsity higher than 95\% without sacrificing generalization performance on testing data. 
	
	However, such sparsity-inducing techniques result in an irregular pattern of sparsity, and require specialized hardware system such as~\cite{han2016eie, chen2018escoin} for speed up. Recent specialized inference systems are typically established by extending well-known Compressed Sparse Row(Column)~(\csrc{}) format~\cite{saad2003iterative} for sparse matrices, and using sparse operations in varying ways. For examples, \eie{}~\cite{han2016eie}  accelerates the fully connected layers of CNN.  Sparse CNN~\cite{liu2015sparse} decomposes the convolutional filters approximately into a set of sparse matrices, (referred as~\scnn{} throughout this paper). \escoin{}~\cite{chen2018escoin} constructs a sequence of sparse matrix multiplications to form convolutional operations on GPUs. 
	
	We have the perspective that \eie, \scnn{} and \escoin{} are valuable state-of-the-art sparse inference systems of CNN with visible advantages and disadvantages. Particularly, \eie{} partially employs sparse operators into CNN, \ie, the fully connected layers, rather than the dominate convolutional layers.~\scnn{} approximates the convolutional filters into sparse matrix multiplications which may result in sacrificing precision of inference. \escoin{} on the other hand, implements the exact sparse convolution by stacking sparse matrix multiplications, but focuses on GPUs, while in the recent world CPUs are still the major inference platform, and GPUs appear more during training stage~\cite{lane2016deepx}. 
	
\begin{figure}[ht!]
		\centering
		\includegraphics[width=\columnwidth]{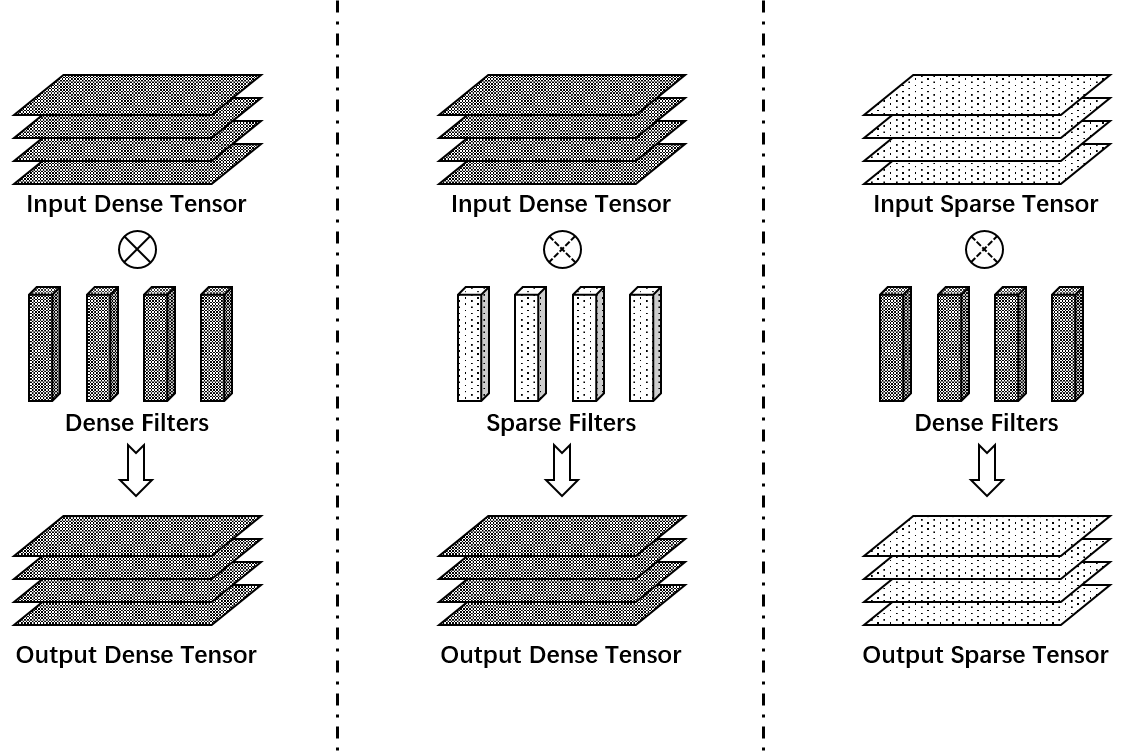}
		\caption{(a): standard convolution operator; (b): \sfalg{}; (c) \sialg{}. }%The convolution operation symbol with dashed lines represents sparse operation.}
		\label{figure:sparse_cnn}
	\end{figure}

	In this paper, we propose, develop, analyze, and provide numerical results for a new sparse convolutional neural network inference accelerator for CPUs, named as~\algname{}~(\algacro). As an overview, our~\algacro{} considers two types of sparsity, \ie, either filters or input tensors of convolutional layers are sparse, tackled by two variants \sfalg{} and \sialg{} respectively, as shown in Figure~\ref{figure:sparse_cnn}. More specifically, in \sfalg{}, we store filters into new designed sparse format, while store input tensor as standard dense format. Similarly, in~\sialg{}, input tensor is represented as sparse format while filters are in dense format.  When either input tensor or filters is high sparse, the number of float operations (FLOPs) reduces dramatically. 
	On the other hand, modern CPUs typically possess multiple processors access to shared memory. To further accelerate the inference procedure, we make use of OpenMP~\cite{dagum1998openmp} to realize multi-core parallelism. The following bulleted points are used to summarize our key contributions.
	
	\begin{itemize}%[label=(\roman*)]
		\item We present a new accelerated Sparse CNN inference framework concentrating on CPUs. Our framework performs efficiently, supports varying sparse modules, \ie, either input or filter weights as sparse format, and is friendly to multi-processors parallelism. 
		
		\item Instead of~\csrc{} format, we make use of a slightly different sparse format, \ie, \liblinear{} sparse format~\cite{fan2008liblinear}. We extend the~\liblinear{} sparse format to represent three-dimensional (3-D) and four-dimensional 4-D tensor decently, design and develop a set of specialized sparse operators to achieve exact and efficient convolutional operation. 
		
		\item The atomic operation of our framework is the inner product between sparse vector and dense vector, different from the matrix multiplication in existing sparse CNN inference frameworks, which typically require additional procedures to construct such operations. Besides, we present algorithms to optimize other related operations in this framework, \eg fusing pooling operation into convolution operation.
		
	\end{itemize}

	\section{Sparsity in CNN}
	
	In many scientific domains, the concept of tensor typically refers to single or multidimensional array, where both vector and matrix are its special instances. If most of elements in tensor are nonzero, then the tensor is considered dense, otherwise sparse. The sparsity in CNN is largely categorized into two classes: (i) sparse filter tensors, and (ii) sparse input tensor. The former one refers to that the trainable parameters of CNNs are sparse (including plentiful zero elements). Perhaps the most popular technique to generate sparse filters is to augment a penalty term to objective function during training, \eg the $\ell_1$-regularized optimization~\cite{chen2018fast} formulated as 
	\begin{equation}\label{prob.x}
	\minimize{x\in \R^n}\ \{F(x)\ \myeq\ f(x)+\lambda\norm{x}_1\}
	\end{equation}	
	where $f(x)$ is the raw objective function of CNN, $\lambda>0$ is a weighting term. Problem~\eqref{prob.x} can be effectively solved deterministically~\cite{fan2008liblinear, andrew2007scalable,chen2017reduced} or stochastically~\cite{jia2018irda,xiao2010dual, xiao2014proximal,chen2021orthant,chen2020neural} to produce pretty sparse solutions even with sparsity higher than 95\% without sacrificing generalization accuracy on testing data.  The latter one describes the  characteristic in data of interest. In some domains, \eg, handwritten applications, data formed into 3-D tensor with numerous zero elements broadly exists~\cite{Chen2017ACC, ji2019generative}. Therefore, utilizing such properties of sparsity to accelerate inference of CNN has practical realistic values.  
	
	\section{Sparse Data Structure}\label{sec:sparsedatastructure}
	
	The standard dense data structure allocates a contiguous memory space to save all entries in tensors. However, operations using such structures may be slow and inefficient when applied to large sparse tensors as processing, and memory are wasted on the zeroes. To address this issue, sparse data structure is introduced to store only non-zero elements that takes advantage of sparsity to avoid unnecessary operations. 
	
\begin{figure}[ht!]
		\centering
		\includegraphics[width=\columnwidth]{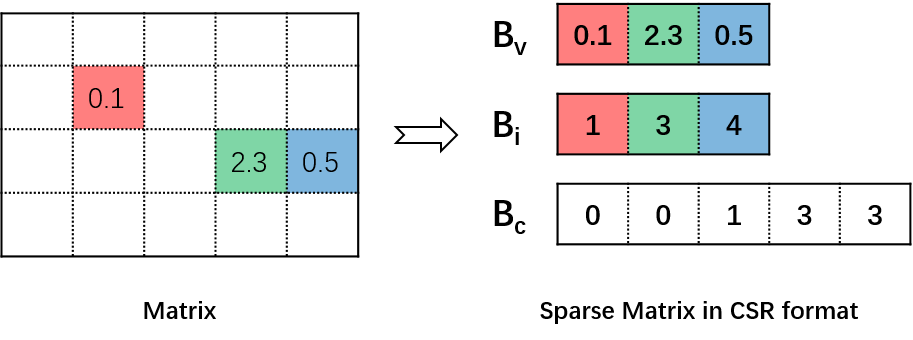}
		\caption{An example of sparse matrix in \csrc{} format.}
		\label{figure:sparse_matrix_csr}
	\end{figure}

	\csrc{} format~\cite{simonyan2014very} is perhaps the most popular sparse format, which represents the sparse matrix by three arrays, as shown in Figure~\ref{figure:sparse_matrix_csr}, (i) $B_v$: all the nonzero entries in this matrix; (ii) $B_i$: column indices of all stored nonzero entries; (iii) $B_c$: accumulated sum of non-zero element quantity (nnz) by each row starting with a 0. But these separate arrays in \csrc{} format may affect memory contiguity, and be not very elegant in real implementation. 

	On the other hand, Liblinear~\cite{fan2008liblinear} implements a slightly different sparse format by utilizing a node struct, (referred as node throughout this paper), to pack values and indices of non-zero elements together, associated with an pointer array to store addresses of starting nodes for each column/row in sparse matrix. Such design is more elegant, and sometimes outperforms standard~\csrc{} format, but is far from sufficiently supporting inference of CNN, which requires extension to 3-D tensor and 4-D tensor. To achieve sparse inference of CNN, we describe basic sparse vector and matrix data structure from~\liblinear{}, and present our extended sparse 3-D and 4-D tensor data structures in the following.
	
	\begin{itemize}[wide, labelwidth=!, labelindent=0pt]
		
		\item \textbf{Node}: Node is the fundamental data structure in~\algacro{}, consisting of two attributes, \ie index and value, to mainly represent non-zero entries in sparse tensors, and some auxiliary components. 
		
		\item \textbf{Sparse Vector}: A sparse vector can be naturally represented by an array of nodes for its non-zero entries with an auxiliary ending node with index as -1, shown as Figure~\ref{figure:sparse_vector}.
		\begin{figure}[ht!]
			\centering
			\includegraphics[width=\columnwidth]{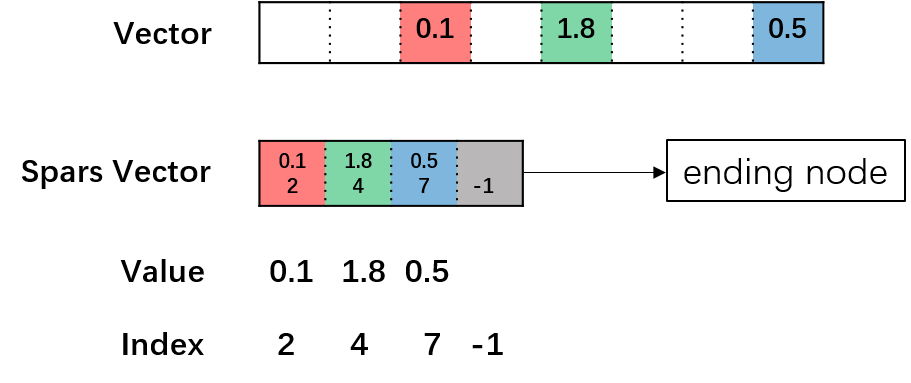}
			\caption{An example of sparse vector.}
			\label{figure:sparse_vector}
		\end{figure}
		\item \textbf{Sparse Matrix}: A sparse matrix can be stored as a list of sparse vectors based on some prescribed order. For matrices, there exist two permissible orders, \ie, $O_{wh}$ and $O_{hw}$, corresponding to saving nodes along first each row (column) then each column (row). Besides, to facilitate access to a specific row (column), additional pointer array is established to store the addresses of the starting nodes of each row(column), illustrated in Figure~\ref{figure:sparse_matrix}. 
		
		\begin{figure}[ht!]
			\centering
			\includegraphics[width=\columnwidth]{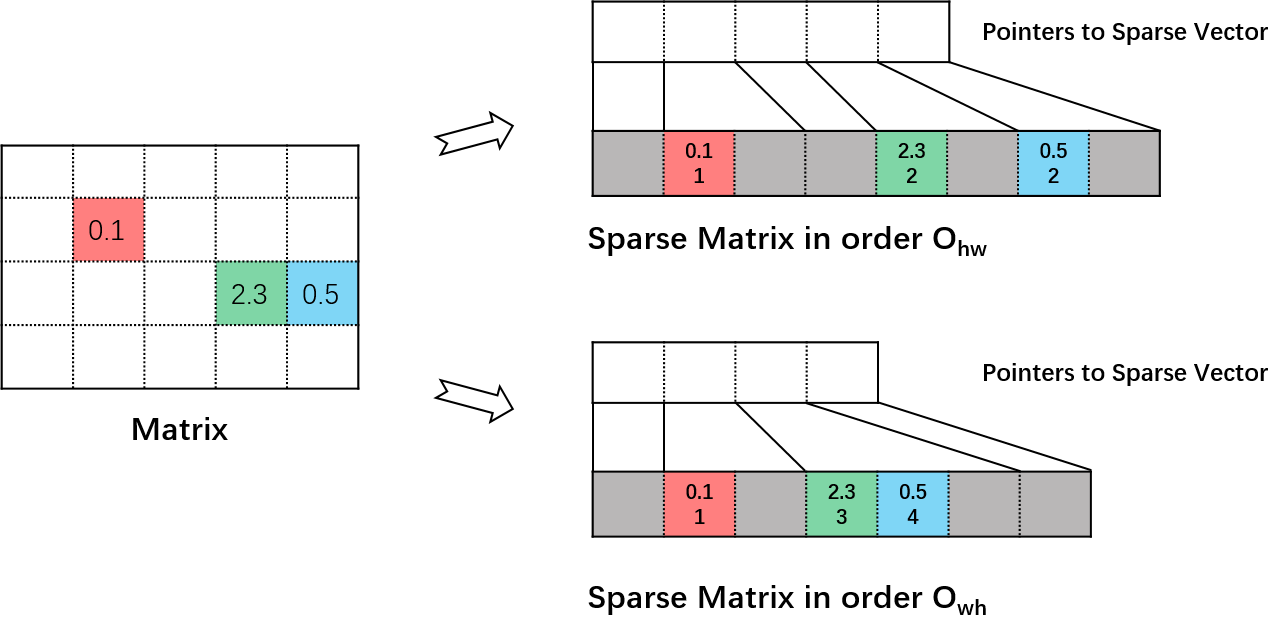}
			\caption{An example of sparse matrix. Blank and gray squares in matrix represent zero elements and ending nodes respectively. }
			\label{figure:sparse_matrix}
		\end{figure}
		
		\item \textbf{Sparse 3-D Tensor}: A sparse 3-D tensor can be viewed as a list of sparse matrices based on some predefined order, which represents the filters or input/output tensors in~\algacro. For 3-D tensors, totally six $(3!=6)$ feasible orders exist, namely, $O_{whc}, O_{wch}, O_{hwc}, O_{hcw}, O_{chw}, O_{cwh}$, in which nodes are saved along specific axis order. For example, in $O_{chw}$, we save nodes along each channel first, height then and width finally. Similarly, to efficient access specific sparse matrix, we make use of an extra pointer array to save the starting node addresses of each matrix, shown in Figure~\ref{figure:sparse_tensor}.
		\begin{figure}[ht!]
			\centering
			\includegraphics[width=\columnwidth]{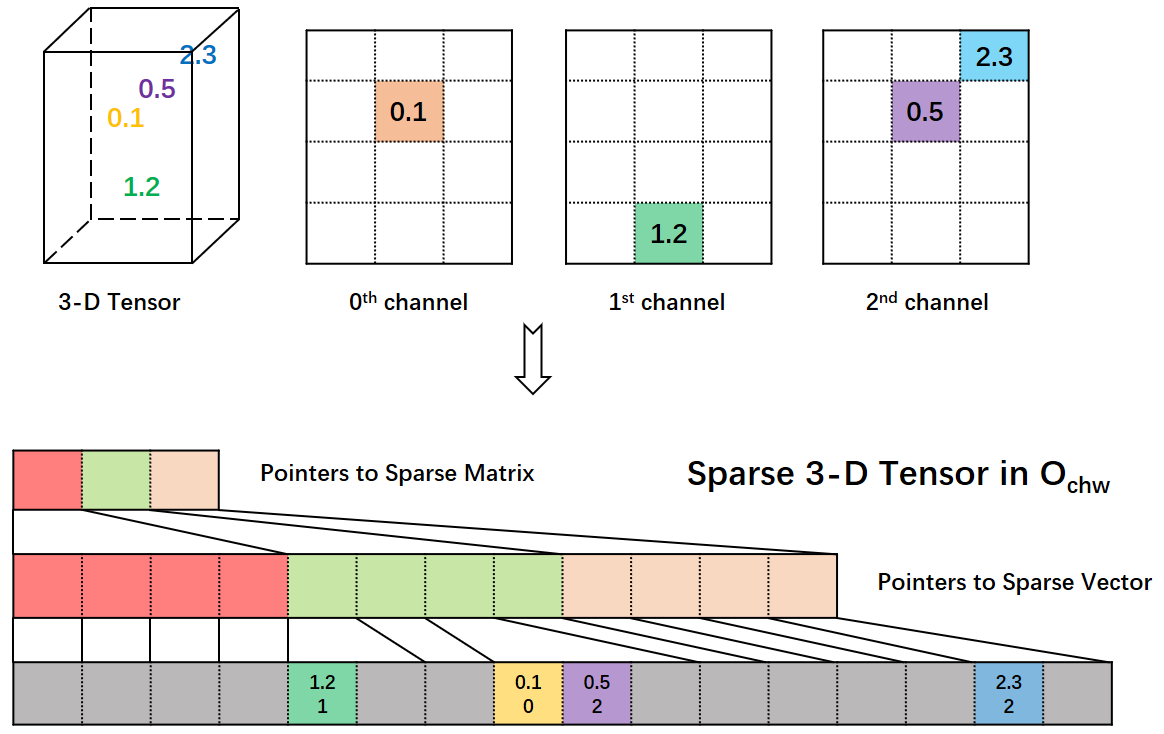}
			\caption{An example of sparse 3-D tensor. }
			\label{figure:sparse_tensor}
		\end{figure}
		
		\item \textbf{Sparse 4-D Tensor}: In~\algacro{}, sparse 4-D tensors are usually interpreted as a batch of input 3-D tensors, or a set of filters in convolutional layers. We make use of the similar logic to store sparse 4-D tensor as a list of sparse 3-D tensor under some order associated with a pointer array for starting nodes. 
		
	\end{itemize}

	\section{Sparse Tensor Operator}\label{sec:sparseoperation}
	
	In this section, we present fundamental sparse tensor operators to form the entire inference of~\algacro{} based on the building blocks described in Section~\ref{sec:sparsedatastructure}. The manner of convolutional operator is first presented. Then we show how to transpose sparse matrix, followed by an extension of operating sparse 3-D tensor transpose.

	\begin{itemize}[wide, labelwidth=!, labelindent=0pt]
		
		\item \textbf{Sparse Tensor Convolution}: Convolution is the most fundamental operation in CNN, which operates on an input 3-D tensor and a 3-D filter tensor to yield a matrix. In~\algacro{}, we pay special interest to the case that either input tensor or filter tensor is sparse corresponding to \sialg{} and~\sfalg{} respectively. To develop efficient sparse convolution, the manner in which we store 3-D tensors matters a lot. After several implementations, we reach a conclusion that perhaps the order $O_{chw}$ is the best, consistent with \pytorch{}~\cite{paszke2017automatic}. We now state the sparse convolution as Algorithm~\ref{alg:sparseconv} for the case that input tensor is dense, and filter tensor is sparse. 
		\begin{algorithm}[ht]
			\caption{Sparse Convolution between Dense Input 3-D Tensor $T_I$ and Sparse Filter Tensor $T_F$.}
			\label{alg:sparseconv}
			\scriptsize
			%\normalsize
			\begin{algorithmic}[1]
				\State \textbf{Input}: $T_I\in\mathbb{R}^{C_I\times H_I\times W_I}$, $T_F\in\mathbb{R}^{C_F\times H_F\times W_F}$, stride $S\in\mathbb{Z}^+$.
				\State \textbf{Output}: A matrix $M$ in dense format.   
				\State Compute number of rows and columns of $M$ as $n_h$ and $n_w$: \label{alg:tensorconv:outputsize}
				\begin{align}
				n_h&\gets (H_I - H_F)/ S + 1;\\
				n_w&\gets (W_I - W_F)/ S + 1.
				\end{align}\label{line:conv_compute_matrix_size}
				\State Declare a dense matrix $M$ of size $(n_h, n_w)$.\label{line:conv_declare_matrix}
				%\State Set $d\gets 0$.\label{alg:tensorconv:d0}
				\For {$i = 0,1,\cdots, n_h-1$}
				\For {$j = 0,1,\cdots, n_w-1$}
				\State Set $x\gets 0$;\label{line:conv_matrix_element_start}
				\For {$l=0,1,\cdots, W_F-1$}
				\For {$k=0,1,\cdots, H_F-1$}
				\State Update $x\gets x + \text{dot}(T_I\verb|[l+jS][k+iS]|, T_F\verb|[l][k]|)$\label{line:sparse_inner_product}
				
				\ \ \ \ \ \ \ \ \ \ \ \ \ \ \ \ \ by Algorithm~\ref{alg:innerproduct}.
				\EndFor
				\EndFor
				%\State Append new value into $M$, \ie $\verb|M[d]| \gets x$.\label{line:append_value_to_matrix} \label{line:conv_matrix_element_end}
				\State Append new value into $M$, \ie $\verb|M[i][j]| \gets x$.\label{line:append_value_to_matrix} \label{line:conv_matrix_element_end}
				%\State Update $d\gets d + 1$
				\EndFor
				\EndFor
				\State \textbf{return} Dense Matrix $M$.
			\end{algorithmic}
		\end{algorithm}
		As shown in Algorithm~\ref{alg:sparseconv}, we first compute the size of resulted matrix $M$ as line~\ref{line:conv_compute_matrix_size}-\ref{line:conv_declare_matrix}. Then each element of matrix is computed by summing $W_FH_F$ inner products between sparse vector from sparse filter $T_F$ and dense vector from input tensor $T_I$ , see line~\ref{line:conv_matrix_element_start} to~\ref{line:conv_matrix_element_end} in Algorithm~\ref{alg:sparseconv}. The crucial is the sparse inner product operator in line~\ref{line:sparse_inner_product} by~Algorithm~\ref{alg:innerproduct}. A concrete example of inner product between dense and sparse vector is shown in Figure~\ref{figure:inner_product}, where float multiplications on zero elements are skipped to reduce the FLOPs. Resulting matrix $M$ is stored by $O_{wh}$ by the order of $T_F$ as $O_{chw}$ and Algorithm~\ref{alg:sparseconv}.
		\begin{algorithm}[ht]
			\caption{Inner Product Between Dense Vector $v_1$ and Sparse Vector $v_2$.}
			\label{alg:innerproduct}
			\begin{algorithmic}[1]
				\State \textbf{Input}: Dense vector $v_1$, sparse vector $v_2$.
				\State \textbf{Output}: Real number $x$.
				\State Set $x\gets 0$.
				\While{$v_2.\text{index}\neq -1$}
				\State Update $x\gets x+v_1[v_2.\text{index}] \cdot v_2.\text{value} $.
				\State Move pointer to next adjacent entry of $v_2$.
				\EndWhile
				\State \textbf{return} $x$.
			\end{algorithmic}
		\end{algorithm}
		
		\begin{figure}[ht!]
			\centering
			\includegraphics[width=0.6\columnwidth]{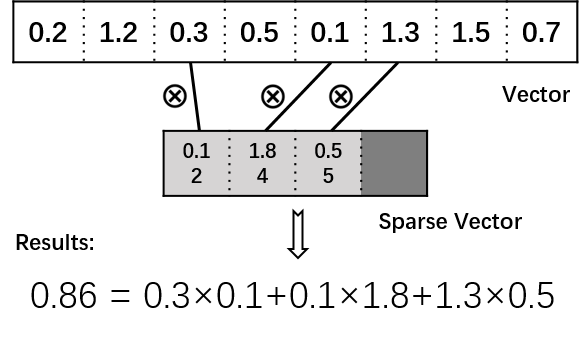}
			\caption{An example of inner product between dense and sparse vector.}
			\label{figure:inner_product}
		\end{figure}
		For the case that input tensor $T_I$ is sparse and filter $T_F$ is dense, the logic of Algorithm~\ref{alg:sparseconv} can be easily extended except setting the format of resulting matrix $M$. In this case, $M$  is stored in sparse format, which requires to check the value of $x$ on line~\ref{line:append_value_to_matrix}, then append only nodes for non-zero $x$'s and auxiliary ending nodes for each row.

		\item \textbf{Sparse Matrix Transpose}: Sparse matrix transpose refers to switch a sparse matrix from order $O_{wh}$ to $O_{hw}$, or reversely.  For simplicity,  transpose from $O_{wh}$ to $O_{hw}$ is described in Algorithm~\ref{alg:transposematrix}, with a concrete example illustrated in Figure~\ref{figure:matrix_transpose}. At first, The indices of the starting point for each column in the new order $O_{hw}$ is calculated as $v$, which is also further used as progress tracker in the remaining transpose operation. Next, the nodes are rearranged from order of $O_{wh}$ into the order of $O_{hw}$ following the progress status shown in $v$. Meanwhile, $v$ is updated if any node is rearranged. The transpose operation completes by setting ending nodes for each column.
		
		\begin{algorithm}[ht]
			\scriptsize
			\caption{Sparse Matrix Transpose from $O_{wh}$ to $O_{hw}$}
			\label{alg:transposematrix}
			\begin{algorithmic}[1]
				\State \textbf{Input}: Sparse matrix $M\in\mathbb{R}^{n_w\times n_h}$ in order $O_{wh}$.
				\State \textbf{Output}: Sparse matrix $M'$ in order $O_{hw}$.
				\State Set $v\in\mathbb{Z}^{n_w+1}$ such that $v[i]$ is the number of nodes (including ending node) in $M$ by the $i$th column. 
				\State Declare $M'$ with appropriate space. 
				\For{$i = 0, 1,\cdots, n_h-1$}
				\For{non-ending node in $i$-th row of $M$}
				\State Set 
				\begin{equation}
				\begin{split}
				M'[v[\text{node.index}]]\text{.index}&\gets i\\
				M'[v[\text{node.index}]]\text{.value}&\gets \text{node.value}.
				\end{split}
				\end{equation}
				\State Update $v[\text{node.index}]\gets v[\text{node.index}]+1$
				\EndFor
				\EndFor
				\For{$i = 0, 1,\cdots, n_w-1$}
				\State Set ending node:
				\begin{equation}
				M'[v[i]]\text{.index} \gets -1
				\end{equation}
				\EndFor
				\State \textbf{return} $M'$.
			\end{algorithmic}
		\end{algorithm}
		
		\begin{figure}[ht!]
			\centering
			\includegraphics[width=0.5\columnwidth]{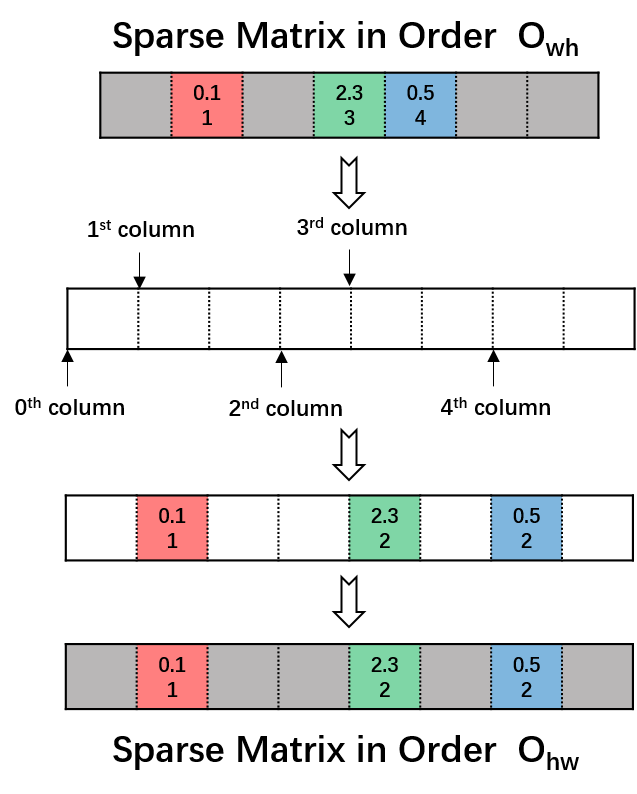}
			\caption{An example of sparse matrix transpose from $O_{wh}$ to $O_{hw}$. The matrix is from Figure~\ref{figure:sparse_matrix}. }
			\label{figure:matrix_transpose}
		\end{figure}
		
		\item \textbf{Sparse 3-D Tensor Transpose}: Similarly to sparse matrix transpose, sparse 3-D tensor transpose refers to convert one sparse 3-D tensor from one order to another. Algorithm~\ref{alg:transposetensor} provides for a sparse 3-D tensor transpose operation, which transposes a 3-D tensor from order $O_{whc}$ to order $O_{chw}$. By the definition of sparse tensor and sparse matrix data structure, a tensor in order $O_{whc}$ can be regarded as number of channels’ sparse matrices in order $O_{wh}$ stacked together. Therefore, the sparse matrix transpose algorithm shown by Algorithm~\ref{alg:transposematrix} is used to transpose each sparse matrix from order $O_{wh}$ to order $O_{hw}$ to form a tensor in $O_{hwc}$. Then, each node is iterated in $O_{hwc}$, reordering them if necessary to obtain a desired sparse tensor in order $O_{chw}$. 
		
		\begin{algorithm}[ht]
			\scriptsize
			\caption{Sparse Tensor Transpose from $O_{whc}$ to $O_{chw}$}
			\label{alg:transposetensor}
			\begin{algorithmic}[1]
				\State \textbf{Input}: Sparse 3-D tensor $T\in\mathbb{R}^{W\times H\times C}$ in order $O_{whc}$.
				\State \textbf{Output}: Sparse 3-D tensor $T'$ in order $O_{chw}$.
				%\Procedure{Sparse Tensor $ T_{O_{chw}}=$ transpose\_tensor}{$T_{O_{whc}}$}
				%\State Set $n_w$, $n_h$, $n_c$ as the width, height, and channel of $T_{O_{chw}}$ respectively.
				\State Declare $T'$ with appropriate space. 
				\For {each sparse matrix $M$ of order $O_{wh}$ in $T$ }
				\State Transpose $M$ from $O_{wh}$ to $O_{hw}$ by Algorithm~\ref{alg:transposematrix}.
				\EndFor
				\For {$w = 0, 1, \cdots, W-1$}
				\For {$h = 0, 1, \cdots, H-1$}
				\For {$c = 0, 1, \cdots, C-1$}
				\If { $w$th column of $c$th $M$ reach non-ending node at $h$th height}\label{line:transposetensor:if}
				\State Copy this node into $T'$ with index as $c$.
				\EndIf
				\EndFor
				\EndFor
				\EndFor
				\State \textbf{return} $T'$.
				%\EndProcedure
			\end{algorithmic}
		\end{algorithm}
	\end{itemize}

	\section{Forward Pass of~\algacro{}}\label{sec:sparse_cnn_forward_pass}
	
	In this section, we assemble the basic data structures and operators described in previous sections to establish the entire forward pass of \cnn{}. A standard~\cnn{} typically consists of stacking varied layers, including convolutional layers with(out) zero-padding, pooling layers, batch normalization layers~\cite{ioffe2015batch}, activation layers, among which convolutional layer is the crucial. Other layers can either operate independently or integrate sequentially into the convolutional layer. 
	Our~\algacro{} considers the later one to avoid unnecessary memory access. Next, we describe the implementations of the above briefly based on the building blocks in Section~\ref{sec:sparsedatastructure} and~\ref{sec:sparseoperation}.
	
	\begin{itemize}[wide, labelwidth=!, labelindent=0pt]
		\item \textbf{Convolutional layer}: Each convolution layer contains an input tensor $T_I\in \mathbb{R}^{C_I\times H_I\times W_I}$, an output tensor $T_O\in \mathbb{R}^{C_O\times H_O\times W_O}$, and a set of filters $T_{F_i}\in \mathbb{R}^{C_{F_i}\times H_{F_i}\times W_{F_ii}}$ ($i\in[1,N]$ , where $N$ is the number of filters). These 3-D tensors satisfy the following properties, (i) $C_I=C_{F_i}$, for all $i\in[1,N]$, (ii) $C_O=N$.  In~\sfalg{} variant, filter tensors $T_{F_i}$ are saved in sparse format, while in~\sialg{} variant, input tensor $T_I$ is saved in sparse format. To implement the forward pass of convolutional layer,~\algacro{} designs two approaches with different advantages, which are illustrated by~\sfalg{} variant as follows.
		
		The first approach is stated as Algorithm~\ref{alg:forwardconvlayer1}. At first, input tensor $T_I$ convolves with each filter to produce $N$ dense matrices of order $O_{wh}$ as line~\ref{line:convTK1}. Then these $N$ dense matrices are stacked along channel axis to form a dense tensor in the order of $O_{whc}$ as line~\ref{line:stackmatrices}. Next we transpose this dense tensor from order of $O_{whc}$ to $O_{chw}$ by Algorithm~\ref{alg:transposetensor}, and return it as the output volume $T_O$ finally. 
		\begin{algorithm}[ht]
			\scriptsize
			\caption{Forward Pass of Convolutional Layer I} 
			\label{alg:forwardconvlayer1}
			\begin{algorithmic}[1]
				\State \textbf{Input}: $T_I\in\mathbb{R}^{C_I\times H_I\times W_I}$, $T_{F_1}, T_{F_2}, \cdots, T_{F_N}\in\mathbb{R}^{C_F\times H_F\times W_F}$, stride $S\in\mathbb{Z}^+$.
				\State \textbf{Output}: Dense 3-D tensor $T_O$ in order $O_{chw}$.
				%\Procedure{Dense $T_{O}=$ foward\_pass\_convlayer}{$T_{I}, T_{K_1}, T_{K_2}, \cdots, T_{K_N}, S$}
				\For {$i=1,2,\cdots, N$} \label{line:forwardconvlayer1:for}
				\State Do convolution operation between dense $T_I$ and sparse $T_{F_i}$ by Algorithm~\ref{alg:sparseconv}.\label{line:convTK1}
				$$
				\text{Dense } M_i\gets \text{tensor\_convolution}(T_I, T_{F_i}, S),
				$$
				\ \ \ \ \ \ where $M_i\in \mathbb{R}^{W_O\times H_O}$ is a dense matrix. 
				\EndFor
				\State Stack $M_i$'s together to form a dense tensor $T_{O'}\in \mathbb{R}^{W_O\times H_O\times N}$. \label{line:stackmatrices}
				\State Transpose $T_{O'}$ from order $O_{whc}$ into order $O_{chw}$ by Algorithm~\ref{alg:transposetensor}.
				$$
				T_O\gets \text{transpose\_tensor}(T_{O'})
				$$
				\State \textbf{return} $T_O$.
				%\EndProcedure
			\end{algorithmic}
		\end{algorithm}
		The second approach stated as Algorithm~\ref{alg:forwardconvlayer2} computes an output tensor more straightforward. Instead of convolving the input tensor with each filter iteratively to obtain multiple matrices, this approach directly compute the elements in output tensor.
		
		The first approach is friendly to multiprocessing on filter level over  line~\ref{line:forwardconvlayer1:for} in Algorithm~\ref{alg:forwardconvlayer1}. Then the loop can be easily split into $N$ independent blocks, and executed by different processors simultaneously. However, the first method creates a temporary list save the output matrices, and processes additional steps of stacking matrices then doing tensor transpose to form output tensor . Hence it is both time and space consuming somewhat. On the other hand, although the second approach skips these extra procedures, its operations are entangled so that its loops can not be easily separated into multiple independent blocks for multiprocessing. This weakness may be overcome when performs inference on a batch of input tensors, where independent blocks over input tensors can be established. 
		
		\begin{algorithm}[ht]
			\scriptsize
			\caption{Forward Pass of Convolutional Layer II}
			\label{alg:forwardconvlayer2}
			\begin{algorithmic}[1]
				\State \textbf{Input}: $T_I\in\mathbb{R}^{C_I\times H_I\times W_I}$, $T_{F_1}, T_{F_2}, \cdots, T_{F_N}\in\mathbb{R}^{C_F\times H_F\times W_F}$, stride $S\in\mathbb{Z}^+$.
				\State \textbf{Output}: Dense 3-D tensor $T_O$ in order $O_{chw}$.
				%\Procedure{Dense $ T_O=$ forward\_pass\_convlayer}{$T_{I}, T_{K_1}, T_{K_2}, \cdots, T_{K_N}, S$}
				\State Calculate $T_O$'s width $W_O$, height $H_O$ and channel $C_O$.\label{alg:forwardconvlayer2:outputsize}
				\begin{align}
				W_O&\gets (W_I - W_{F_1})/ S + 1;\\
				H_O&\gets (H_I - H_{F_1})/ S + 1;\\
				C_O&\gets N.
				\end{align}
				\State Declare a dense 3-D tensor $T_O$ of size $(W_O, H_O, C_O)$.\label{line:conv_declare_matrix}
				%\State Set $d\gets 0$.\label{alg:forwardconvlayer2:d0}
				%\State Set $r\gets 0$.\label{alg:forwardconvlayer2:r0}
				\For {$w = 0,1,\cdots, W_O-1$}
				%\State Update: $T_O.\verb|dataSS[w]|\gets T_O.\verb|dataS[r]|$
				%\State Set $T_O.\verb|dataSS[w]|$ as the address of $T_O.\verb|dataS[r]|$
				\For {$h = 0,1,\cdots, H_O-1$}
				%\State Update: $T_O.\verb|dataS[r]|\gets T_O.\verb|data[d]|$
				%\State Set $T_O.\verb|dataS[r]|$ as the address of $T_O.\verb|data[d]|$
				%\State Update: $r\gets r+1$
				\For {$f=1,2,\cdots, C_O$}
				\State Set $x\gets 0$;
				\For {$l=0,1,\cdots, W_F-1$}
				\For {$k=0,1,\cdots, H_F-1$}
				\State Update: $
				x\gets x + \text{dot}(T_I\verb|[l+wS][k+hS]|, T_{F_{f}}\verb|[l][k]|)
				$\label{alg:forwardconvlayer2:dot}
				\EndFor
				\EndFor
				%\If{ $x\neq 0$} \label{line:check-x-zero}
				%\State Append new value into $T_O$, \ie $\verb|M[d]| \gets x$.\label{line:append_value_to_matrix} \label{line:conv_matrix_element_end}
				%\State Append new value into $T_O$:
				%\State $$
				%T_O.\verb|data[d] = x|
				%$$
				\State Append new value into $T_O$, \ie, $T_O\verb|[w][h][f-1]| \gets x$.
				%\State Update $d\gets d + 1$
				%\EndIf
				\EndFor
				%\State Add ending node into $T_O$:
				%\State $$\verb|T.data[d].index| = -1$$ \label{alg:forwardconvlayer2:endfor}
				
				\EndFor
				\EndFor
				\State \textbf{return} Dense Tensor $T_O$.
				%\EndProcedure
			\end{algorithmic}
		\end{algorithm}
		
		Our~\algacro{} contains a switch between these approaches to capitalize on the their advantages while avoiding the disadvantages. In particular, Algorithm~\ref{alg:forwardconvlayer2} is called by default, unless the input batch size is relative small then Algorithm~\ref{alg:forwardconvlayer1} is performed.
		
		\item \textbf{Pooling Layers}:  Straightforward way to implement pooling layer is to perform pooling operation directly on the output tensor from previous layer. However, in \sialg{} variant, it is difficult to locate the pooling region for sparse tensor of  $O_{chw}$ because of its specific storage order. Therefore, for \sialg{}, we merge pooling operations with convolutional operations stated as Algorithm~\ref{alg:mergepoolconv}, which is more efficient than doing convolutional and pooling operation separately. At first, the resulting matrix size is calculated using the parameters of convolutional and pooling operators. Then, convolutional operations are performed, as described above. However, during this operation, when an element is generated, it is used to update an array used for tracking values for each pooling region. When the pooling regions are complete, a sequence of values is constructed from the tracking value array. A matrix is generated in a sparse matrix data structure in order $O_{wh}$ by merging convolutional and pooling operations. An example of merged pooling operation is shown in Figure~\ref{figure:mergepoolconv}. As for \sfalg{} variant, the above merging mechanism also works well, but we found out that doing convolutional and pooling operation separately performs more efficiently.

		\begin{algorithm}[ht]
			\scriptsize
			\caption{Merge Pooling and Convolutional Operation} 
			\label{alg:mergepoolconv}
			\begin{algorithmic}[1]
				\State \textbf{Input}: $T_I\in\mathbb{R}^{C_I\times H_I\times W_I}$, $T_{F}\in\mathbb{R}^{C_F\times H_F\times W_F}$, stride $S\in\mathbb{Z}^+$, pooling operator $P$ of size ($H_P$, $W_P$).
				\State \textbf{Output}: A matrix $M$ in dense format.
				
				%\Procedure{Dense Matrix $M=$ tensor\_convolution\_pooling}{$T_1,T_2, S, P$}
				\State Calculate the size of resulted matrix by only convolutional operation:
				\begin{align}
				n_{h,conv}&\gets (H_I - H_F)/ S + 1;\\
				n_{w,conv}&\gets (W_I - W_F)/ S + 1.   
				\end{align}
				\State Calculate $M$'s number of rows $n_r$ and columns $n_c$ after both convolutional and pooling operation.\label{alg:tensorconvpool:outputsize}
				\begin{align}
				n_{h, pool}&\gets n_{h, conv}/H_P;\\
				n_{w, pool}&\gets n_{w, conv}/W_P.
				\end{align}
				%where \verb|P| denotes the pooling operator, \verb|P.height| is the pooling operator's height, and \verb|P.width| is pooling operator's width. 
				%where \verb|P| denotes the pooling operator, $H_P$ is the pooling operator's height, and $W_P$ is pooling operator's width. 
				\State Declare \verb|poolTrackValues| as a size of $n_{w, pool}$ array to track desired values for each pooling region with size $H_P\times W_P$. 
				%\State Declare \verb|poolTrackIndexes| as a size of $n_{c, pool}$ array to track desired value's index for each pooling region with size $\verb|P.height|\times \verb|P.width|$. 
				\For {$i = 0,1,\cdots, n_{h,conv}-1$}
				\For {$j = 0,1,\cdots, n_{w,conv}-1$}
				\State Set $x\gets 0$;
				\For {$l=0,1,\cdots, W_F-1$}
				\For {$k=0,1,\cdots, H_F-1$}
				\State Update:
				$x\gets x + \text{dot}(T_I\verb|[l+jS][k+iS]|, T_F\verb|[l][k]|)$
				\EndFor
				\EndFor
				%\State Update \verb|poolTrackValues[Pooling region of x]| by $x$ if needed. 
				\State Update $$\verb|poolTrackValues[Pooling region of x]|$$ by $x$ if needed. 
				%\State Update \verb|poolTrackIndexes[Pooling region of x]| by $j$ if needed.\\
				
				\EndFor
				%\If{All calculations in current pooling region is complete }
				\If{$(i+1) \% H_P == 0$}
				%\State Obtain values from \verb|poolTrackValues|.
				%\State Insert these values into $M$.
				\State Insert values from \verb|poolTrackValues| into $M$.
				\State Reset \verb|poolTrackValues|.
				\EndIf
				\EndFor
				\State \textbf{return} $M$.
				%\EndProcedure
			\end{algorithmic}
		\end{algorithm}
		
		\begin{figure}[ht!]
			\centering
			\includegraphics[width=\columnwidth]{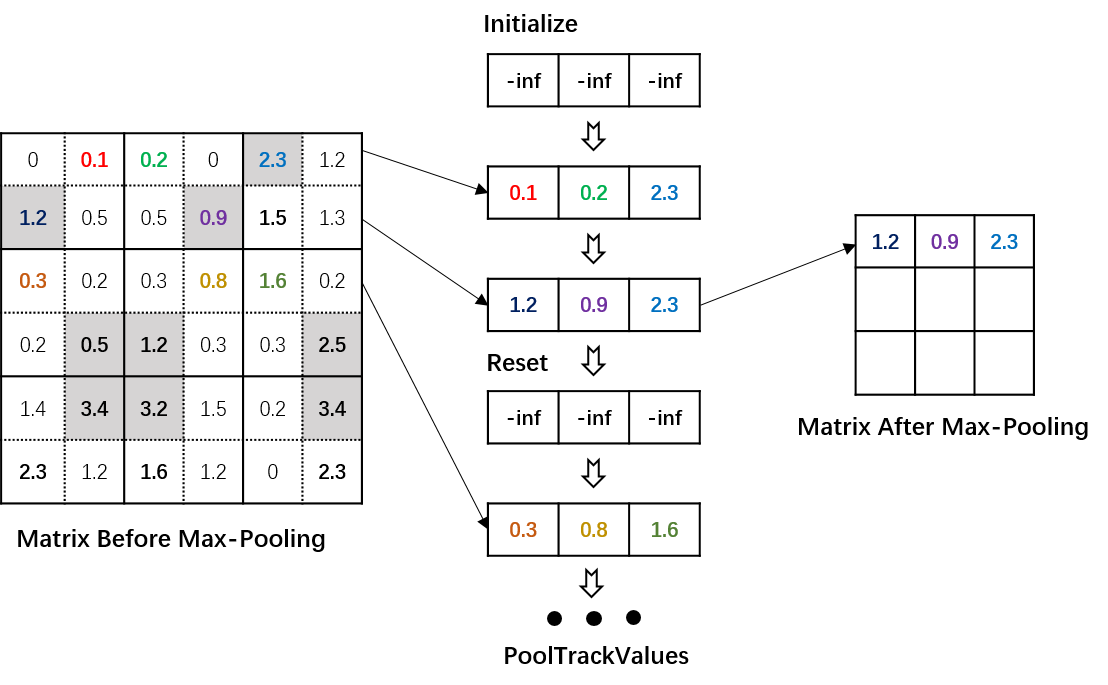}
			\caption{An example of merged convolution and max-pooling operation as described in Algorithm~\ref{alg:mergepoolconv}. The bold number above gray grid represents the maximum entry in the pooling region. }
			\label{figure:mergepoolconv}
		\end{figure}

		\item \textbf{Other layers}: \algacro{} is flexible to padding operation, batch normalization and activation operations. Padding layers can be implemented by setting the output matrix with the same shape as the input tensor and shifting the indexing of input tensor in inner product left or right. Both batch normalizatiton and activation layer are additional functions applied to each elements in output tensor. % Activation functions, such as Rectified Linear Unit(ReLU)~\cite{nair2010rectified}, might introduce zero elements in output tensor, so they should be revoked before we append new node to the output tensor in \sialg{}.
		
		\item \textbf{Forward pass of~\algacro{}}: 	The forward pass of sparse CNN becomes pertty straightforward once the above fundamental layers are well-defined.  In general, layers are serially performed, and the output tensor of current layer serves as the input tensor of next layer. 
		
	\end{itemize}

	\section{Numerical Experiments}\label{sec.numexp}
	In this section, we  numerically demonstrate the efficiency of~\algacro{} by comparing it with the state-of-the-art deep learning library \pytorch{} on popular CNN architectures~\vgg{}~\cite{simonyan2014very} and~\yolo{}~\cite{redmon2016you}. All experiments are conducted on a 64-bit machine with one Intel Xeon Platinum 8163 CPU with 16 processors and 32GB of main memory. \pytorch{} version is $0.4.1$ without GPU support. To be consistent with the default data type in \pytorch{}, our implementation uses 32-bit float numbers rather than 64-bit double numbers.
	
	As described in Section~\ref{sec:sparse_cnn_forward_pass}, when the batch size of inputs is no greater than a threshold,~\algacro{} performs Algorithm~\ref{alg:forwardconvlayer1}, otherwise triggers Algorithm~\ref{alg:forwardconvlayer2}. This threshold is determined by the physical environment. In our experiments, we set it as $4$. We make use of \openmp{}~\cite{dagum1998openmp} to achieve multiprocessing programming. Generally, \openmp{} exactly reduces the runtime of \algacro{} in most cases, but when available CPU resource is limited, the cost of revoking \openmp{} overcomes the benefits of multiprocessing. To achieve best performance, we turn off the \openmp{} if CPU usage is larger than $0.95$ for \sfalg{}.

	\subsection{\sfalg{} Experiments}
	To illustrate the performance of \sfalg{} comprehensively, we evaluate runtimes of \sfalg{} for input as single or multiple instances under filter tensors of varying densities.

	\begin{itemize}[wide, labelwidth=!, labelindent=0pt]
		
		\item \textbf{Single Instance}: We compare the performance between~\algacro{} and~\pytorch{} on~\vgg{} and~\yolo, shown in Table~\ref{table:mode2_vgg} and Table~\ref{table:mode2_yolo} respectively.  It can be observed that~\algacro{} generally performs better than \pytorch{} when the density is no greater than $5\%$, indicating that \sfalg{} effectively takes the advantage of high sparsity to accelerate inference. Especially when the density is less than $1\%$, \sfalg{} outperforms \pytorch{} regardless of number of processes, and can achieve 1.5x to 3x speedups. Following the increase of density,~\algacro{} becomes slower since the benefits of sparse operations gradually disappear.

		\begin{table}
			\label{table:mode2}
			\begin{subtable}[t]{\columnwidth}
				\centering
				\resizebox{\columnwidth}{!}{
					\def\arraystretch{1.1}
					\begin{tabular}{cccccccc}
						\Xhline{3\arrayrulewidth}
						\multirow{2}{*}{density(\%)} & \multirow{2}{*}{methods} & \multicolumn{6}{c}{number of processes running simultaneously}  \\
						\cline{3-8}
						& & 1 & 2 & 4 & 8 & 12 & 16 \\
						\hline
						\multirow{2}{*}{1} & \textbf{\algacro{}} & \cellcolor{Gray}{ 0.1622} & \cellcolor{Gray}{0.2437} & \cellcolor{Gray}{ 0.3992} & \cellcolor{Gray}{ 0.6909} & \cellcolor{Gray}{ 0.9813} & \cellcolor{Gray}{ 1.1903} \\ 
						& \pytorch{} & {\color{black} 0.5542} & {\color{black} 0.5580} & {\color{black} 0.6114} & {\color{black} 0.9007} & {\color{black} 1.3887} & {\color{black} 1.8918}  \\ 
						\hline
						\multirow{2}{*}{5} & \textbf{\algacro{}} & \cellcolor{Gray}{ 0.3067} & \cellcolor{Gray}{ 0.3907} & {\color{black} 0.6394} & {\color{black} 1.1156} & {\color{black} 1.5435} & \cellcolor{Gray}{ 1.7810} \\ 
						& \pytorch{} & {\color{black} 0.5531} & {\color{black} 0.5728} & {\color{black} 0.6001} & {\color{black} 0.8964} & {\color{black} 1.3864} & {\color{black} 1.8892}  \\ 
						\hline
						\multirow{2}{*}{10} & \textbf{\algacro{}} & {\color{black} 0.6029} & {\color{black} 0.5934} & {\color{black} 0.9693} & {\color{black} 1.7233} & {\color{black} 2.4091} & {\color{black} 2.8139} \\ 
						& \pytorch{} & {\color{black} 0.5408} & {\color{black} 0.5638} & {\color{black} 0.5997} & {\color{black} 0.8870} & {\color{black} 1.3793} & {\color{black} 1.8875} \\ 
						\hline
						\multirow{2}{*}{20} & \textbf{\algacro{}} & {\color{black} 1.3414} & {\color{black} 0.9872} & {\color{black} 1.6748} & {\color{black} 3.0375} & {\color{black} 4.2691} & {\color{black} 5.1231} \\ 
						& \pytorch{} & {\color{black} 0.5418} & {\color{black} 0.5644} & {\color{black} 0.5963} & {\color{black} 0.8886} & {\color{black} 1.3749} & {\color{black} 1.8885} \\ 
						\hline
						\multirow{2}{*}{50} & \textbf{\algacro{}} & 4.6124 & {\color{black} 2.1707} & {\color{black} 3.7652} & {\color{black} 7.0721} & {\color{black} 9.8160} & {\color{black} 12.0788} \\ 
						& \pytorch{} & {\color{black} 0.5402} & {\color{black} 0.5645} & {\color{black} 0.5985} & {\color{black} 0.8841} & {\color{black} 1.3576} & {\color{black} 1.8860}  \\ 
						\hline
						\multirow{2}{*}{100} & \textbf{\algacro{}} & 9.7058 & 4.1137 & 7.2583 & 13.8396 & 18.9156 & 23.4767  \\ 
						& \pytorch{} & {\color{black} 0.5463} & {\color{black} 0.5664} & {\color{black} 0.6005} & {\color{black} 0.8874} & {\color{black} 1.3369} & {\color{black} 1.8944}  \\ 
						\Xhline{3\arrayrulewidth}
					\end{tabular}
				}
				\caption{Runtime for \vgg{}.\label{table:mode2_vgg}}
			\end{subtable}
			\newline
			\vspace*{0.5 cm}
			\newline
			\begin{subtable}[t]{\columnwidth}
				\centering
				\resizebox{\columnwidth}{!}{
					\def\arraystretch{1.1}
					\begin{tabular}{cccccccc}
						\Xhline{3\arrayrulewidth}
						\multirow{2}{*}{density(\%)} & \multirow{2}{*}{methods} & \multicolumn{6}{c}{number of processes running  simultaneously} \\
						\cline{3-8}
						& & 1 & 2 & 4 & 8 & 12 & 16 \\
						\hline
						\multirow{2}{*}{1} & \textbf{\algacro{}} & \cellcolor{Gray}{ 0.1803} & \cellcolor{Gray}{ 0.2978} & \cellcolor{Gray}{ 0.4577} & \cellcolor{Gray}{ 0.7315} & \cellcolor{Gray}{ 0.8765} & \cellcolor{Gray}{ 0.9503} \\
						& \pytorch{} & {\color{black} 0.4884} & {\color{black} 0.5261} & {\color{black} 0.6230} & {\color{black} 1.2258} & {\color{black} 2.0018} & {\color{black} 2.7346} \\
						\hline
						\multirow{2}{*}{5} & \textbf{\algacro{}} & \cellcolor{Gray}{ 0.3912} & \cellcolor{Gray}{ 0.4583} & {\color{black} 0.6969} & \cellcolor{Gray}{ 1.1374} & \cellcolor{Gray}{ 1.4834} & \cellcolor{Gray}{ 1.6194} \\
						& \pytorch{} & {\color{black} 0.4721} & {\color{black} 0.5280} & {\color{black} 0.6130} & {\color{black} 1.2173} & {\color{black} 1.9816} & {\color{black} 2.7419} \\
						\hline
						\multirow{2}{*}{10} & \textbf{\algacro{}} & {\color{black} 0.7359} & {\color{black} 0.6182} & {\color{black} 0.9794} & {\color{black} 1.6397} & {\color{black} 2.2129} & \cellcolor{Gray}{ 2.4725} \\
						& \pytorch{} & {\color{black} 0.4925} & {\color{black} 0.5197} & {\color{black} 0.6047} & {\color{black} 1.2097} & {\color{black} 1.9690} & {\color{black} 2.7209} \\
						\hline
						\multirow{2}{*}{20} & \textbf{\algacro{}} & 1.4992 & {\color{black} 0.9536} & {\color{black} 1.5086} & {\color{black} 2.6650} & {\color{black} 3.6570} & {\color{black} 4.2331} \\
						& \pytorch{} & {\color{black} 0.4922} & {\color{black} 0.5140} & {\color{black} 0.6186} & {\color{black} 1.2135} & {\color{black} 1.9669} & {\color{black} 2.7250} \\
						\hline
						\multirow{2}{*}{50} & \textbf{\algacro{}} & 4.1250 & {\color{black} 1.8676} & {\color{black} 3.1060} & {\color{black} 5.8067} & {\color{black} 7.9827} & {\color{black} 9.6181} \\
						& \pytorch{} & {\color{black} 0.4941} & {\color{black} 0.5123} & {\color{black} 0.6056} & {\color{black} 1.2082} & {\color{black} 1.9175} & {\color{black} 2.7212} \\
						\hline
						\multirow{2}{*}{100} & \textbf{\algacro{}} & 8.8167 & 3.4057 & 5.7335 & 11.1056 & 15.1332 & {\color{black} 18.5831} \\
						& \pytorch{} & {\color{black} 0.4956} & {\color{black} 0.5258} & {\color{black} 0.6126} & {\color{black} 1.2250} & {\color{black} 1.8401} & {\color{black} 2.7353} \\
						
						\Xhline{3\arrayrulewidth}
					\end{tabular}
				}
				\caption{Runtime for \yolo{}.\label{table:mode2_yolo}}
			\end{subtable}
			\caption{Runtimes of \algacro{} and \pytorch{}  under the architecture of (a) \vgg{} (b) \yolo{} with different densities and number of processes running simultaneously. ~\algacro{} is set as~\sfalg{} variant. }
		\end{table}

		\item \textbf{Multiple Instances}: To explore the performance of \sfalg{} given a batch of input tensors, we create Table \ref{table:mode3_vgg} and \ref{table:mode3_yolo} to present the runtimes on \vgg{} and \yolo{} respectively.
		We can see that when the density is no larger than $5\%$, \sfalg{} runs faster than other approach with one exception, \ie, the test of density as $5\%$ with batch size being 4. Even if density increases up to $20\%$, \sfalg{} still achieves faster inference than \pytorch{} on \yolo{} when batch size equals to $16$ or $32$. This verifies the efficiency of underlying forward pass in Algorithm~\ref{alg:forwardconvlayer2}.  Particularly, when density equals to $1\%$ and  batch size is larger than $8$, \sfalg{} can achieves roughly 6x speedups on both \vgg{} and \yolo{}.

		\begin{table}
			\label{table:mode3}
			\begin{subtable}[t]{\columnwidth}
				\centering
				\resizebox{\columnwidth}{!}{
					\def\arraystretch{1.1}
					\begin{tabular}{ccccccccc}
						\Xhline{3\arrayrulewidth}
						\multirow{2}{*}{density(\%)} & \multirow{2}{*}{methods} & \multicolumn{7}{c}{batch size} \\
						\cline{3-9}
						& & 1 & 2 & 4 & 8 & 16 & 32 & 64 \\
						\hline
						\multirow{2}{*}{1} & \textbf{\algacro{}} & \cellcolor{Gray}{ 0.1394} & \cellcolor{Gray}{ 0.3124} & \cellcolor{Gray}{ 0.6214} & \cellcolor{Gray}{ 0.7828} & \cellcolor{Gray}{ 1.1650} & \cellcolor{Gray}{ 2.2829} & \cellcolor{Gray}{ 4.5582} \\
						& \pytorch{} & {\color{black} 0.5413} & {\color{black} 0.9982} & {\color{black} 1.8448} & {\color{black} 3.4790} & {\color{black} 6.7911} & {\color{black} 13.5070} & {\color{black} 26.7536} \\
						\hline
						\multirow{2}{*}{5} & \textbf{\algacro{}} & \cellcolor{Gray}{ 0.3159} & \cellcolor{Gray}{ 0.6134} & \cellcolor{Gray}{ 1.0934} & \cellcolor{Gray}{ 1.6125} & \cellcolor{Gray}{ 2.5824} & \cellcolor{Gray}{ 5.1019} & \cellcolor{Gray}{ 10.2115} \\
						& \pytorch{} & {\color{black} 0.5442} & {\color{black} 0.9813} & {\color{black} 1.8453} & {\color{black} 3.4805} & {\color{black} 6.8337} & {\color{black} 13.5011} & {\color{black} 26.7553} \\
						\hline
						\multirow{2}{*}{10} & \textbf{\algacro{}} & {\color{black} 0.5979} & {\color{black} 1.1789} & {\color{black} 2.1062} & \cellcolor{Gray}{ 2.6579} & \cellcolor{Gray}{ 4.0500} & \cellcolor{Gray}{ 8.0308} & \cellcolor{Gray}{ 16.1110} \\
						& \pytorch{} & {\color{black} 0.5427} & {\color{black} 0.9962} & {\color{black} 1.8509} & {\color{black} 3.4892} & {\color{black} 6.7933} & {\color{black} 13.4856} & {\color{black} 26.7686} \\
						\hline
						\multirow{2}{*}{20} & \textbf{\algacro{}} & {\color{black} 1.3436} & {\color{black} 2.6965} & {\color{black} 3.6095} & {\color{black} 4.8044} & \cellcolor{Gray}{ 6.7654} & \cellcolor{Gray}{ 13.5081} & {\color{black} 27.0243} \\
						& \pytorch{} & {\color{black} 0.5428} & {\color{black} 0.9594} & {\color{black} 1.8275} & {\color{black} 3.5040} & {\color{black} 6.8366} & {\color{black} 13.5211} & {\color{black} 26.6894} \\
						\hline
						\multirow{2}{*}{50} & \textbf{\algacro{}} & 4.4832 & 9.3022 & 14.9400 & {\color{black} 11.3700} & {\color{black} 13.7979} & {\color{black} 27.5494} & {\color{black} 55.0434} \\
						& \pytorch{} & {\color{black} 0.5410} & {\color{black} 0.9876} & {\color{black} 1.8414} & {\color{black} 3.4889} & {\color{black} 6.8329} & {\color{black} 13.4898} & {\color{black} 26.7946} \\
						\hline
						\multirow{2}{*}{100} & \textbf{\algacro{}} & 9.7428 & 19.4605 & 38.2397 & 22.2444 & 24.7809 & 49.6027 & 99.3588 \\
						& \pytorch{} & {\color{black} 0.5517} & {\color{black} 1.0005} & {\color{black} 1.8418} & {\color{black} 3.4857} & {\color{black} 6.8532} & {\color{black} 13.5203} & {\color{black} 26.6328} \\

						\Xhline{3\arrayrulewidth}
					\end{tabular}
				}
				\caption{Runtime for \vgg{}.\label{table:mode3_vgg}}
			\end{subtable}
			\newline
			\vspace*{0.5 cm}
			\newline
			\begin{subtable}[t]{\columnwidth}
				\centering
				\resizebox{\columnwidth}{!}{
					\def\arraystretch{1.1}
					\begin{tabular}{ccccccccc}
						\Xhline{3\arrayrulewidth}
						\multirow{2}{*}{density(\%)} & \multirow{2}{*}{methods} & \multicolumn{7}{c}{batch size} \\
						\cline{3-9}
						& & 1 & 2 & 4 & 8 & 16 & 32 & 64 \\
						\hline
						\multirow{2}{*}{1} & \textbf{\algacro{}} & \cellcolor{Gray}{ 0.1785} & \cellcolor{Gray}{ 0.3820} & \cellcolor{Gray}{ 0.7946} & \cellcolor{Gray}{ 0.7581} & \cellcolor{Gray}{ 1.1605} & \cellcolor{Gray}{ 2.3004} & \cellcolor{Gray}{ 4.6041} \\
						& \pytorch{} & {\color{black} 0.4582} & {\color{black} 0.8680} & {\color{black} 1.6231} & {\color{black} 2.9958} & {\color{black} 5.7639} & {\color{black} 11.3392} & {\color{black} 23.0086} \\
						\hline
						\multirow{2}{*}{5} & \textbf{\algacro{}} & \cellcolor{Gray}{ 0.3959} & \cellcolor{Gray}{ 0.8375} & {\color{black} 1.7294} & \cellcolor{Gray}{ 1.4390} & \cellcolor{Gray}{ 2.1726} & \cellcolor{Gray}{ 4.3269} & \cellcolor{Gray}{ 8.7502} \\
						& \pytorch{} & {\color{black} 0.4869} & {\color{black} 0.8716} & {\color{black} 1.6311} & {\color{black} 3.0023} & {\color{black} 5.7948} & {\color{black} 11.4490} & {\color{black} 22.9784} \\
						\hline
						\multirow{2}{*}{10} & \textbf{\algacro{}} & {\color{black} 0.7364} & {\color{black} 1.5338} & {\color{black} 3.0927} & \cellcolor{Gray}{ 2.3183} & \cellcolor{Gray}{ 3.2735} & \cellcolor{Gray}{ 6.5320} & \cellcolor{Gray}{ 13.0444} \\
						& \pytorch{} & {\color{black} 0.4888} & {\color{black} 0.8696} & {\color{black} 1.6326} & {\color{black} 3.0095} & {\color{black} 5.7911} & {\color{black} 11.4212} & {\color{black} 23.0659} \\
						\hline
						\multirow{2}{*}{20} & \textbf{\algacro{}} & 1.4927 & 3.0760 & 6.2619 & {\color{black} 4.0802} & \cellcolor{Gray}{ 5.2731} & \cellcolor{Gray}{ 10.5313} & \cellcolor{Gray}{ 21.0404} \\
						& \pytorch{} & {\color{black} 0.4907} & {\color{black} 0.8714} & {\color{black} 1.6355} & {\color{black} 3.0035} & {\color{black} 5.8353} & {\color{black} 11.4885} & {\color{black} 22.9948} \\
						\hline
						\multirow{2}{*}{50} & \textbf{\algacro{}} & 4.1997 & 8.1943 & 16.7630 & {\color{black} 9.7099} & {\color{black} 11.3508} & {\color{black} 22.6907} & {\color{black} 45.3361} \\
						& \pytorch{} & {\color{black} 0.4883} & {\color{black} 0.8663} & {\color{black} 1.6524} & {\color{black} 3.0169} & {\color{black} 5.8191} & {\color{black} 11.4623} & {\color{black} 22.9997} \\
						\hline
						\multirow{2}{*}{100} & \textbf{\algacro{}} & 8.8026 & 17.9287 & 34.9626 & 19.4020 & 21.2747 & 42.5343 & 85.1374 \\
						& \pytorch{} & {\color{black} 0.4900} & {\color{black} 0.8759} & {\color{black} 1.6315} & {\color{black} 3.0023} & {\color{black} 5.8114} & {\color{black} 11.4506} & {\color{black} 23.0047} \\
						\Xhline{3\arrayrulewidth}
					\end{tabular}
				}
				\caption{Runtime for \yolo{}.\label{table:mode3_yolo}}	
			\end{subtable}
			\caption{Runtimes of \algacro{} and \pytorch{} under the architecture of (a) \vgg{} (b) \yolo{} with different batch sizes. ~\algacro{} is set as~\sfalg{} variant.}
		\end{table}
		
	\end{itemize}

	\subsection{\sialg{} Experiments}
	
	In this section, we investigate the performance of \sialg{} variant which assumes  input tensor is sparse and filters are dense. To receive significant computational benefits from sparsity, the input tensor typically supposed to be highly sparse.  Similar to the \sfalg{} experiments, we select the density candidates as $0.01$, $0.05$, $0.1$, $0.2$, $0.5$ and $1$ of input tensor to evaluate the performance of \sialg{}. 
	
	\begin{table}[h]
		\setlength{\arrayrulewidth}{.07em}
		\centering
		\resizebox{\columnwidth}{!}{
			\begin{tabular}{|c|c||cccccc|}
				\hline
				\multirow{2}{*}{model} & \multirow{2}{*}{method} & \multicolumn{6}{c|}{density (\%)} \\
				\cline{3-8}
				& & 1 & 5 & 10 & 20 & 50 & 100 \\
				\Xhline{2\arrayrulewidth}
				\multirow{2}{*}{\vgg{}} & \textbf{\algacro{}} &  1.1582 & 1.3453  & 1.3748 & 1.3873 & 1.3923 & 1.3721 \\
				& \pytorch{} &  0.5412 & 0.5423  & 0.5454 & 0.5443 & 0.5449 & 0.5439 \\
				\hline
				\multirow{2}{*}{\yolo{}} & \textbf{\algacro{}} &  1.5754 & 1.5833  & 1.5952 & 1.6061 & 1.6241 & 1.5885 \\
				& \pytorch{} &  0.4820 & 0.4685  & 0.4707 & 0.4971 & 0.4691 & 0.4692 \\
				\hline
			\end{tabular}
		}
		\caption{Runtime of~\algacro{} and~\pytorch{} on \vgg{} and \yolo{}.~\algacro{} is set as~\sialg{}.}
		\label{table:sialg_all}
	\end{table}
	
	Table~\ref{table:sialg_all} shows the runtimes of \sialg{} and \pytorch{} under different densities, where~\pytorch{} performs faster than~\algacro{} on every test of~\vgg{} and~\yolo{}. There are a few reasons as follows:
	\begin{itemize}[wide, labelwidth=!, labelindent=0pt]
		
		\item \vgg{} and~\yolo{} consists of stacking a large number of convolutional layers. Even though the input tensor of the first convolutional layer is assumed as sparse, the sparsity of intermediate layers decays dramatically because of the convolution operation. Therefore, the gain from sparse operations vanishes gradually during forward pass. For a clear illustration, we create Figure~\ref{figure:sialg_density_all} to present the evolution of densities of input tensors over each convolutional layer for \vgg{} and \yolo{}, on which the input tensors generally becomes denser and denser. 
		
		\begin{figure}%[h]
			\begin{subfigure}[t]{0.235\textwidth}
				\centering
				\includegraphics[width=\textwidth]{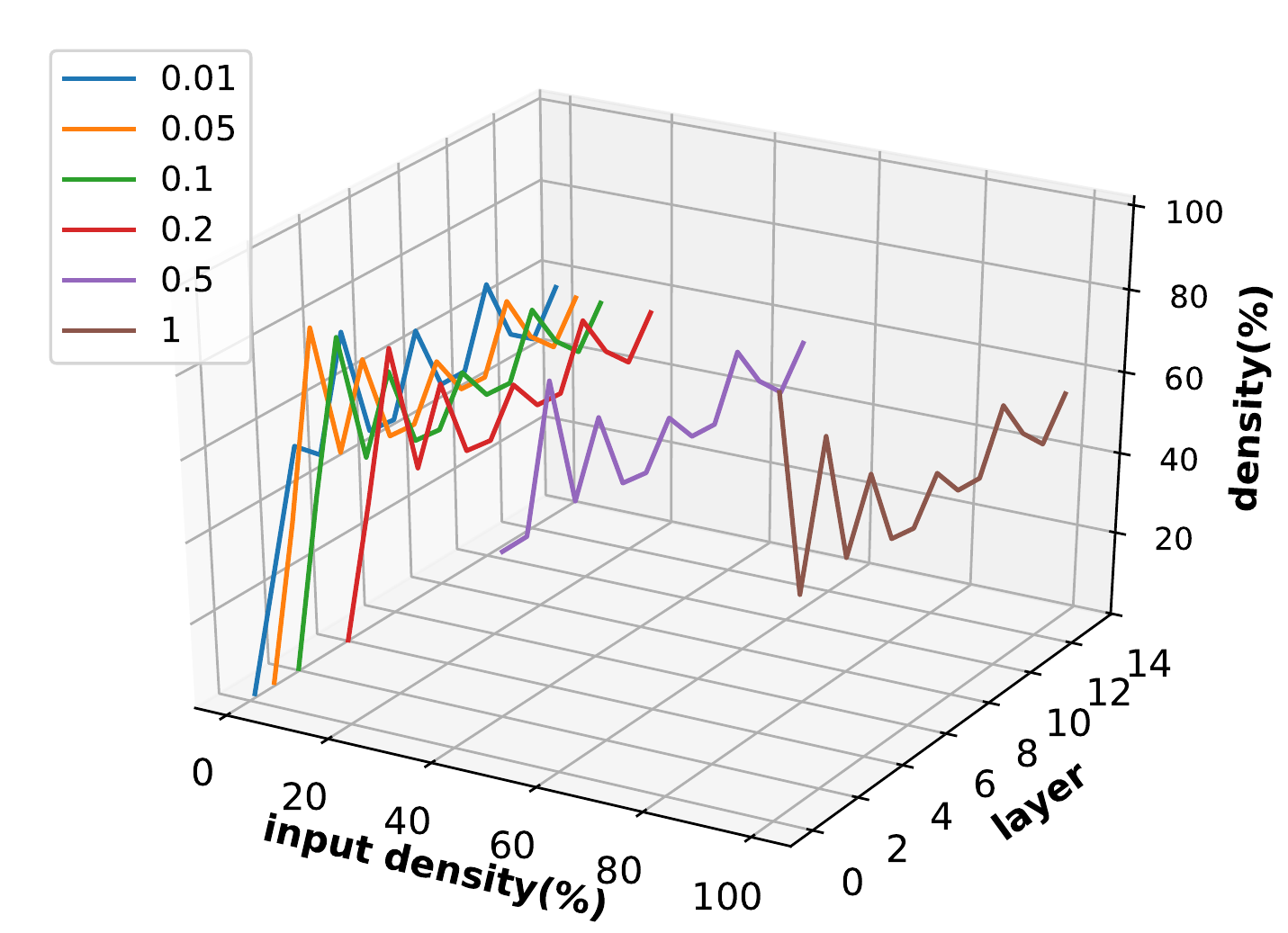}
				\caption{\vgg{}}
				\label{figure:sialg_vgg16_all}
			\end{subfigure}
			\begin{subfigure}[t]{0.235\textwidth}
				\centering
				\includegraphics[width=\textwidth]{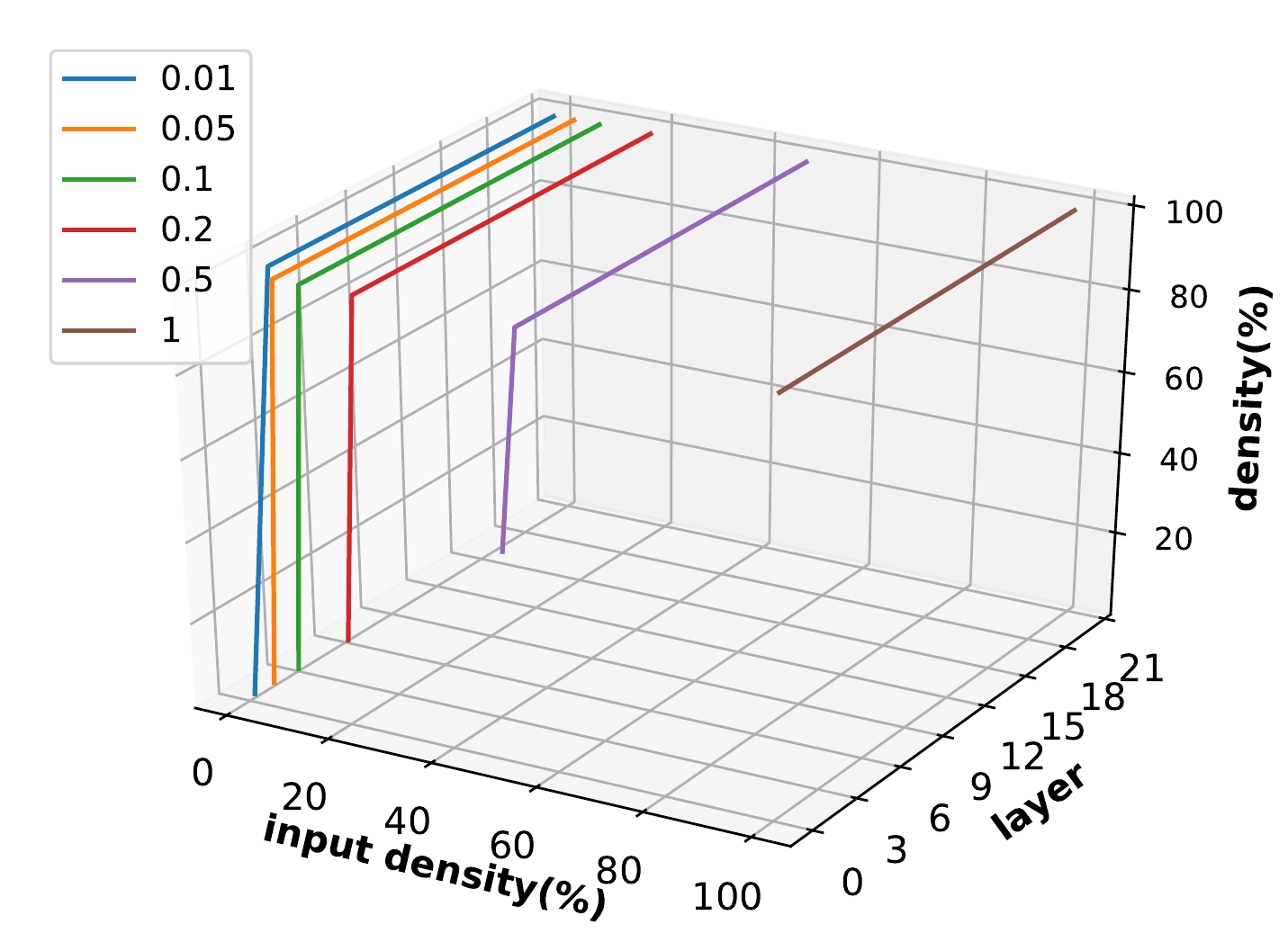}
				\caption{\yolo{}}
				\label{figure:sialg_yolo_all}
			\end{subfigure}
			\caption{Density at each layer for (a) \vgg{}; (b) \yolo{} with different densities of input tensor.}
			\label{figure:sialg_density_all}
		\end{figure}
		
		\item Activation and batch normalization layers also affects the sparsity a lot.  As shown in~Figure~\ref{figure:sialg_vgg16_all}, the density curve of~\vgg{} fluctuates fiercely, because of the existence of ReLu activation function in~\vgg{}. ReLu projects negative entries back to zero, so that the sparsity can be somehow promoted. On the contrast, \yolo{} makes use of LeakyReLU~\cite{maas2013rectifier} and batch normalization, so that the density rapidly increases to $100\%$ only after a few convolutional layers, shown in Figure~\ref{figure:sialg_yolo_all}. 
		
		To verify the above statements, we remove activation functions and batch normalization layers from \vgg{} and~\yolo{}, then record the runtime in Table~\ref{table:sialg_pure} and the density evolution in Figure~\ref{figure:sialg_density_pure}.  \vgg{} without ReLU runs slower than standard \vgg{}, which can be interpreted by the density curves that \vgg{} with ReLU tends to possess sparser input tensors, so that~\sialg{} receives more computational benefits. Therefore, to let \sialg{} achieve the best performance, it is recommended to construct a shallow convolutional neural network stacking by a small number of convolutional layers, and carefully select the activation functions.

		\begin{figure}%[ht!]
			\begin{subfigure}[t]{0.23\textwidth}
				\centering
				\includegraphics[width=\textwidth]{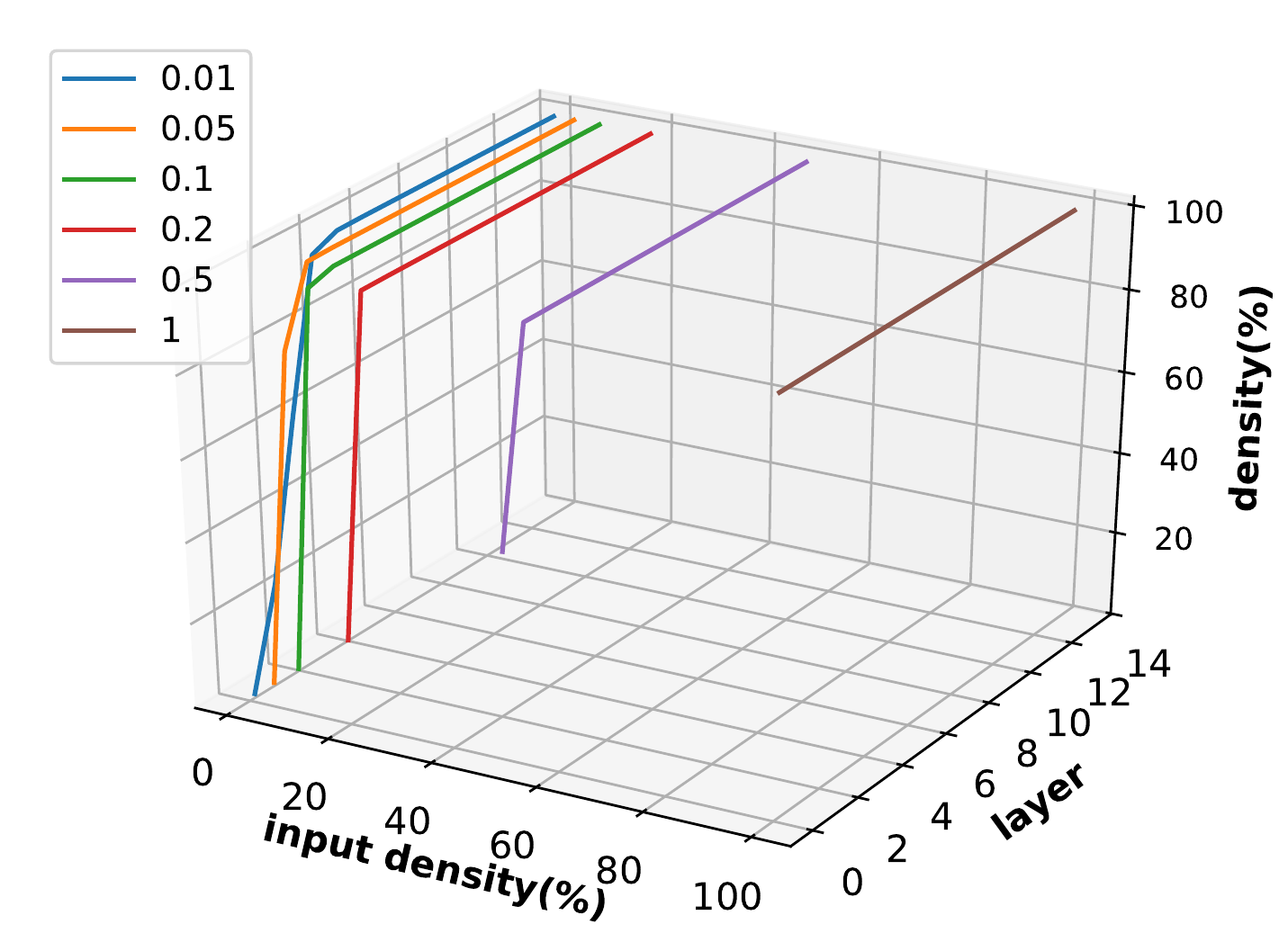}
				\caption{\vgg{}}
				\label{figure:sialg_vgg16_pure}
			\end{subfigure}
			\begin{subfigure}[t]{0.23\textwidth}
				\centering
				\includegraphics[width=\textwidth]{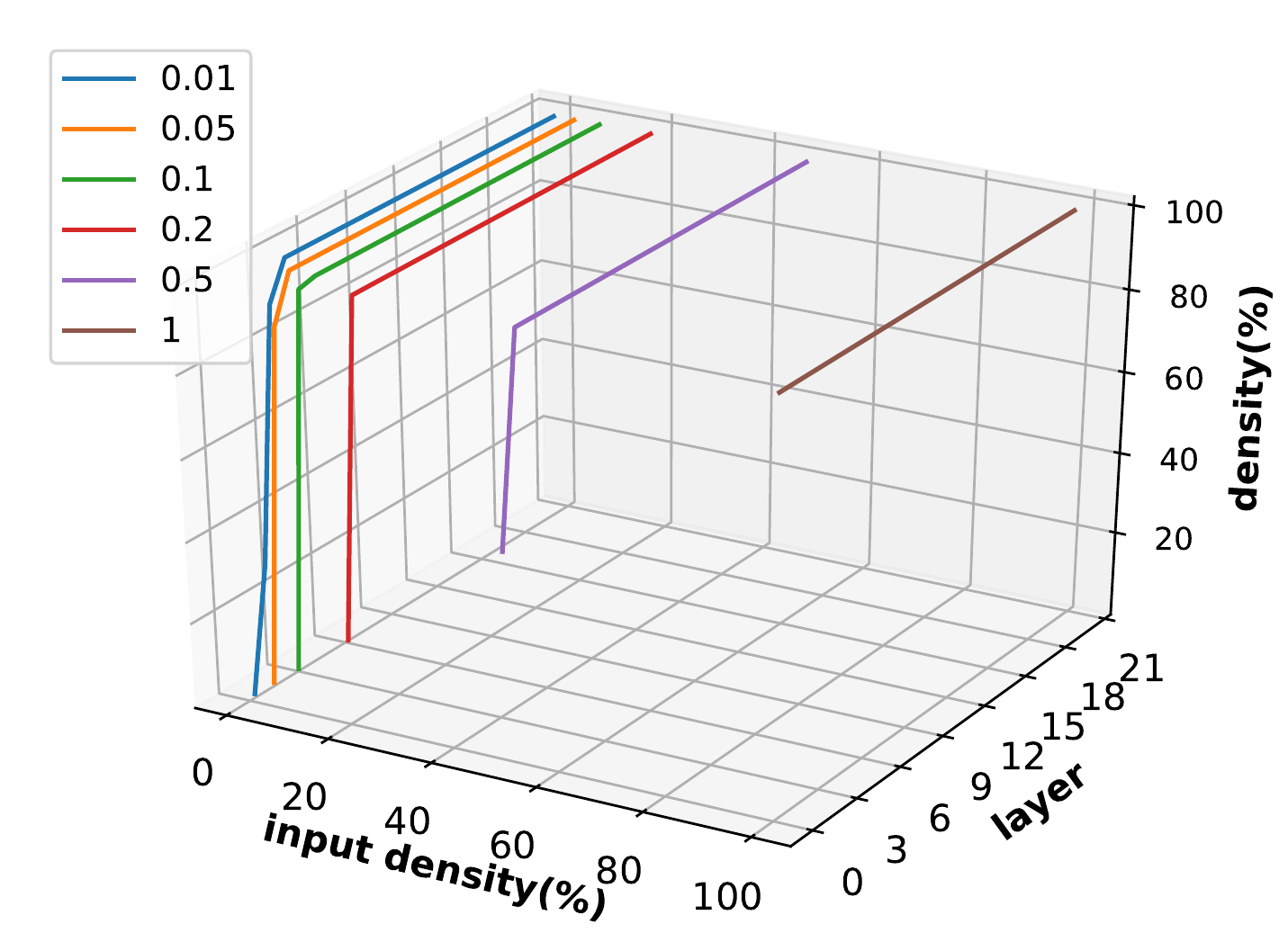}
				\caption{\yolo}
				\label{figure:sialg_yolo_pure}
			\end{subfigure}
			\caption{Density at each layer for (a) \vgg{}; (b) \yolo{} with different densities of input . We remove the activation layers and batch normalizations from \vgg{} and \yolo{}.}
			\label{figure:sialg_density_pure}
		\end{figure}

	\end{itemize}
	
	\begin{table}
		\setlength{\arrayrulewidth}{.07em}
		\centering
		\resizebox{\columnwidth}{!}{
			\begin{tabular}{|c|c||cccccc|}
				\hline
				\multirow{2}{*}{model} & \multirow{2}{*}{method} & \multicolumn{6}{c|}{density} \\
				\cline{3-8}
				& & 1 & 5 & 10 & 20 & 50 & 100 \\
				\Xhline{2\arrayrulewidth}
				\multirow{2}{*}{\shortstack{\vgg{}\\ (-ReLU)}} & \textbf{\algacro{}} &  1.6478 & 1.9160  & 1.9419 & 1.9516 & 1.9638 & 1.9436 \\
				& \pytorch{} &  0.5262 & 0.5308  & 0.4970 & 0.5441 & 0.5288 & 0.5291 \\
				\hline
				\multirow{2}{*}{\shortstack{\yolo{}\\ (-BN, -LeakyReLU)}} & \textbf{\algacro{}} &  1.5957 & 1.6767  & 1.6869 & 1.6941 & 1.7138 & 1.6763 \\
				& \pytorch{} &  0.4358 & 0.4347  & 0.4361 & 0.4356 & 0.4354 & 0.4353 \\
				\hline
			\end{tabular}
		}
		\caption{Density of output tensor and runtime for \sialg{} on \vgg{} and \yolo{}. We remove the activation layers and batch normalizations from \vgg{} and \yolo{}.}
		\label{table:sialg_pure}
	\end{table}

	%*********
	% Section Conclusions
	%*********
	%\section{Conclusion}\label{sec.conclusion}
	
%	We proposed a fast sparse convolution neural network inference system, so called~\algacro{} to accelerate the forward pass of CNN on CPU. To establish it, we designed specialized sparse data structures, operators, and associated algorithms to fit multiprocessing.~\algacro{} is efficient and friendly to parallelism.  It has a framework consisting of two variants, namely~\sfalg{} and~\sialg{}, to consider two types of sparsity, \ie, either filters or input tensors are sparse. Experiments on popular deep convolution neural networks, such as \vgg{} and \yolo{}, showed that \sfalg{} outperforms state-of-the-art deep learning library~\pytorch{} when filters are highly sparse, and revealed the potential of~\sialg{} for sparse input tensor under proper model architecture setting. 

	%*********
	% Section Acknowledgments
	%*********
	%\section*{Acknowledgments}
	%\newpage
	{\small
		\bibliographystyle{ieee_fullname}
		\bibliography{paperbib}
		%\bibliography{bibliography}
	}
	
\end{document}